\documentclass[a4paper,fleqn]{cas-dc}

\usepackage[authoryear,longnamesfirst]{natbib}

\usepackage{amsmath}
\usepackage{soul}

\def\tsc#1{\csdef{#1}{\textsc{\lowercase{#1}}\xspace}}
\tsc{WGM}
\tsc{QE}
\tsc{EP}
\tsc{PMS}
\tsc{BEC}
\tsc{DE}

\begin{document}
\let\WriteBookmarks\relax
\def\floatpagepagefraction{1}
\def\textpagefraction{.001}

\shorttitle{Swarm Intelligence in Biomedical Sciences}

\shortauthors{S. Adhikary}

\title [mode = title]{Nature Inspired Evolutionary Swarm Optimizers for Biomedical Image and Signal Processing - A Systematic Review}                      


%
\author[1]{Subhrangshu Adhikary}[
                        orcid=https://orcid.org/0000-0003-1779-3070]

\cormark[1]


\ead{subhrangshu.adhikary@spiraldevs.com}



\affiliation[1]{organization={Department of Research \& Development, Spiraldevs Automation Industries Pvt. Ltd.},
    city={Raiganj},
    postcode={733123}, 
    state={West Bengal},
    country={India}}

\cortext[cor1]{Corresponding author}

\begin{abstract}
The challenge of finding a global optimum in a solution search space with limited resources and higher accuracy has given rise to several optimization algorithms. Generally, the gradient-based optimizers converge to the global solution very accurately, but they often require a large number of iterations to find the solution. Researchers took inspiration from different natural phenomena and behaviours of many living organisms to develop algorithms that can solve optimization problems much quicker with high accuracy. These algorithms are called nature-inspired meta-heuristic optimization algorithms. These can be used for denoising signals, updating weights in a deep neural network, and many other cases. In the state-of-the-art, there are no systematic reviews available that have discussed the applications of nature-inspired algorithms on biomedical signal processing. The paper solves that gap by discussing the applications of such algorithms in biomedical signal processing and also provides an updated survey of the application of these algorithms in biomedical image processing. The paper reviews 28 latest peer-reviewed relevant articles and 26 nature-inspired algorithms and segregates them into thoroughly explored, lesser explored and unexplored categories intending to help readers understand the reliability and exploration stage of each of these algorithms.
\end{abstract}


\begin{highlights}
\item Latest relevant 28 papers were surveyed regarding the application of nature-inspired algorithms in biomedical signal and image processing.
\item 26 Nature-inspired meta-heuristic algorithms were discussed.
\item Most frequently used, moderately explored and lesser explored nature-inspired algorithms were identified. 
\end{highlights}


\maketitle

\section{Introduction}
\label{sec:intro}
The neural network's weights, many decision support systems and feature extraction methods often require a reliable, fast and robust optimization method for optimum performance. Gradient descent and its derivative algorithms like Adam, Stochastic gradient descent, etc. are very reliable and robust optimizers. However, as these methods perform a thorough search into the solution space, these are quite slow. Therefore there exists a need for faster but reliable optimization algorithms. Researchers over time have observed different natural phenomena and swarming behaviours of various animals, insects, birds, etc. while foraging or hunting. These behaviours are mathematically simulated to solve optimization problems. Different algorithms have their different strengths and weakness but most of them are quite faster than gradient-based algorithms. These algorithms generally consist of two phases, one is an exploration where optima are searched and the other one is exploitation where those searched optima are checked whether it is global or local optima. These algorithms can be used to solve different problems in biomedical signal and image processing. This includes denoising, compression, feature extraction, weight updates, etc. 
In the present state-of-the-art, there are no standard reviews published which discuss the application of nature-inspired algorithms for solving optimization tasks in biomedical signal and image processing and therefore the paper reviews all such works where these algorithms were used to solve optimization problems in the biomedical signal and image processing domain.

\begin{figure*}
    \centering
    \includegraphics[scale=0.6]{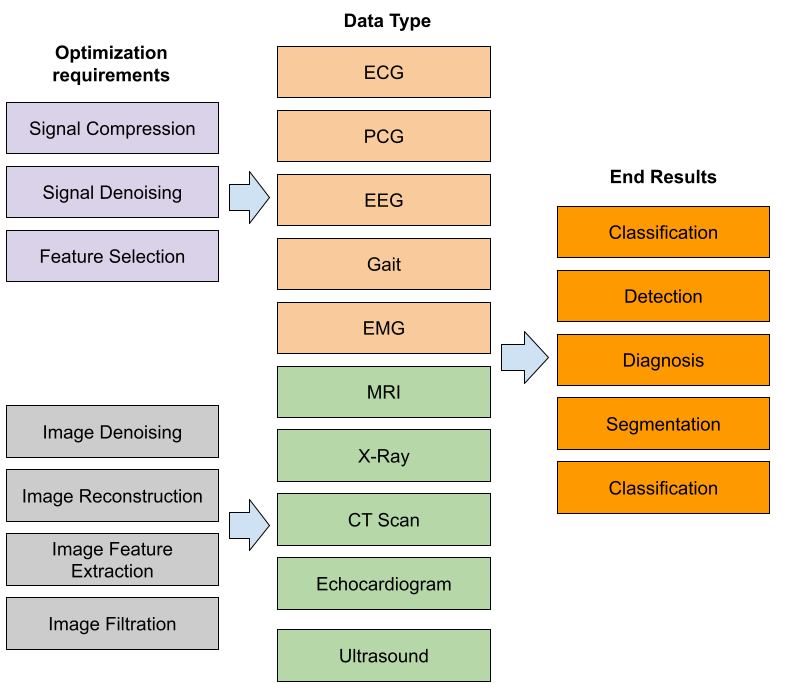}
    \caption{Applications of optimizers in different types of biomedical data and their end results}
    \label{fig:biomedical}
\end{figure*}

In the biomedical engineering discipline, optimization algorithms are required various stages. Figure \ref{fig:biomedical} shows where the optimizers are required in which kind of biomedical data and their end results. Biomedical signals like EEG, EMG, and ECG often contain several noises. These require adaptive filtering techniques to eliminate those. Biomedical images like CT, XRay, MRI, etc. also require adaptive noise reduction. Both of these can be effectively solved by using nature-inspired adaptive filters~\cite{WANG2021376}. Different classification tasks of biomedical signal and image processing intended for use in diagnosis often requires robust feature selection mechanisms. This is because when unwanted features are fed into a classification model, the machine learning decision can be adversely affected. Therefore a reliable feature selection method helps to get a better classification accuracy~\cite{pashaei2022efficient}. Some diagnosis modelling involves the application of image reconstruction techniques to develop an image from biomedical signals obtained from various sensors~\cite{ben2021deep}. Some signals like ECG and some biomedical images can be compressed to save storage space and the compression can be performed without losing the essential features by using evolutionary algorithms~\cite{geetha2021evolutionary}. Hyperparameters of different deep learning models could be fine-tuned and even the weights can be updated by using nature-inspired algorithms~\cite{wei2020review}.

Throughout the literature, different surveys have been conducted to review the current state-of-the-art nature-inspired meta-heuristic algorithms and their application in the biomedical engineering discipline. The summary of the state of the art is given in Table ~\ref{tab:state-of-the-art-reviews}. ~\cite{nayak2020firefly} have reviewed the applications of the firefly algorithm in the healthcare domain. The work is a generalized study and does not focus on signal or image processing. ~\cite{rundo2020survey,MITRA2015111} has further published review articles in which they have surveyed the application of bio-inspired algorithms with biomedical images but these study does not discuss the signal processing applications.  Cross-disciplinary studies like ~\cite{li2020newly,https://doi.org/10.1002/cjce.24246} was published to show recent advanced in the swarm and evolutionary intelligence but it was not focused towards biomedical signal or image processing. A review on adaptive filtering was published by ~\cite{Kumar_Manish_2020} for denoising of biomedical images but signal processing was not considered in this work. These limitations have motivated to conduct a survey to review the usage of the nature-inspired meta-heuristic swarm and evolutionary intelligence methods in solving different optimization problems in biomedical signal and image processing.

\begin{table*}
    \centering
    \caption{State of the Art Surveys related to applications of nature-inspired meta-heuristic optimization algorithms for biomedical image and signal processing}
    \label{tab:state-of-the-art-reviews}
    \begin{tabular}{p{3.5cm}|p{4.5cm}|p{3.5cm}|p{4cm}}
    \hline
        Source & Theme & Objective & Limitations  \\
        \hline
        ~\cite{nayak2020firefly} & Review of firefly algorithm in biomedical sciences and health care & Reviewed 111 papers related to FA in the domain & Generalized study, not focused on signal or image processing \\
        \hline
        ~\cite{rundo2020survey} & Survey of nature-inspired algorithms for biomedical image processing & To study the clinical decision support system for biomedical images using swarm and evolutionary algorithms & Signal processing not discussed \\
        \hline
        ~\cite{li2020newly} & Detailed review of newly emerged nature-inspired algorithms & 17 Algorithms were discussed including their applications & Cross-disciplinary applications reviewed and not specialized to biomedical signal and image processing \\
        \hline
        ~\cite{Kumar_Manish_2020} & Review on adaptive filtering based on nature-inspired neural network & To remove noises in medical images & Application of signal processing with nature-inspired algorithm not focused \\
        \hline
        ~\cite{https://doi.org/10.1002/cjce.24246} & Review of AI-based process control in biochemical and biomedical sciences & 280 paper including applications of nature-inspired algorithms and other approaches were discussed & Cross-disciplinary generalized study, not focused on signal and image processing \\
        \hline
        ~\cite{MITRA2015111} & Survey image processing techniques with the application of nature-inspired algorithms for cancer diagnosis & Usage of swarm intelligence methods on CT, PET and MRI data & Limited discussion on signal processing techniques \\
        \hline
    \end{tabular}
\end{table*}

The primary contribution of the work is:
\begin{itemize}
    \item To study the literature and review all relevant papers published from 2012-2024 that discuss the application of nature-inspired optimizers in biomedical signal and image processing.
    \item The study has categorised the algorithms that are highest explored, moderately explored and lesser explored.
    \item Attempts to find the algorithms which have the potential to provide reliable performance for solving optimization problems but are not enough explored.
    \item To understand the areas in biomedical engineering in which nature-inspired algorithms can be used.
\end{itemize}

In the paper, section \ref{sec:methods} discusses the methods for conducting the survey. Following this, the algorithms hence found were divided into 5 most explored (section \ref{sec:freq-algorithms}), 7 partly explored (section \ref{sec:moderately-algorithms}) and 14 unexplored (section \ref{sec:less-algorithms}) algorithms. Lastly section \ref{sec:conclusion} concludes the paper. 

\section{Methodology}
\label{sec:methods}
The paper presents a systematic review that aims to identify in the literature works related to the application of nature-inspired meta-heuristic optimization algorithms in biomedical image and signal processing. For this purpose, the worldwide trusted, peer-reviewed popular digital libraries and databases were explored namely Scopus, Web of Science, ScienceDirect ACM Digital Library, MDPI, IEEE Xplore Digital Library, Wiley, Springer and Hindawi. A generic search query was used for finding the papers for the review. It is given as: (TITLE-ABS-KEY(nature AND inspired AND algorithm AND biomedical) OR TITLE-ABS-KEY(nature AND inspired AND biomedical AND optimizer) OR TITLE-ABS-KEY(hybrid AND glow AND worm OR swarm AND optimization AND algorithm OR mine AND bomb AND algorithm OR water AND cycle AND algorithm OR cuckoo AND search OR quantum-behaved AND particle AND swarm AND optimization OR design AND by AND shopping AND paradigm OR dolphin AND echolocation AND algorithm OR artificial AND fish AND swarm OR firefly AND algorithm OR invasive AND weed AND optimization OR albatross AND wind AND shear AND algorithm OR cat AND swarm AND optimization OR moth-flame AND optimization AND algorithm OR galactic AND swarm AND optimization OR hummingbirds AND optimization AND algorithm OR flower AND pollination AND algorithm OR artificial AND flora AND optimization AND algorithm OR whale AND optimization AND algorithm OR cockroach AND swarm AND optimization OR grey AND wolf AND optimizer OR whale AND optimization AND algorithm OR biogeography-based AND optimization OR bacterial AND foraging AND optimisation OR bird AND mating AND optimizer OR artificial AND bee AND colony AND algorithm OR fruit AND fly AND optimization AND algorithm OR imperialst AND competitive AND algorithm OR big AND bang AND - AND big AND crunch AND algorithm OR harmony AND search AND algorithm OR ant AND colony AND optimization OR bat AND algorithm OR penguins AND search AND optimization AND algorithm OR intelligent AND water AND drops OR grenade AND explosion AND method OR teaching--learning-based AND optimization OR salp AND swarm AND algorithm OR elephant AND herding AND optimization OR virus AND spread AND optimization)) AND PUBYEAR > 2012 AND PUBYEAR < 2024 AND PUBYEAR > 2013 AND PUBYEAR < 2024 AND ( LIMIT-TO ( DOCTYPE,"ar" ) OR LIMIT-TO ( DOCTYPE,"cp" ) OR LIMIT-TO ( DOCTYPE,"ch" ) ) AND ( LIMIT-TO ( SRCTYPE,"j" ) OR LIMIT-TO ( SRCTYPE,"p" ) ) AND ( LIMIT-TO ( LANGUAGE,"English" ) )

\begin{figure*}
    \centering
    \includegraphics[scale=0.6]{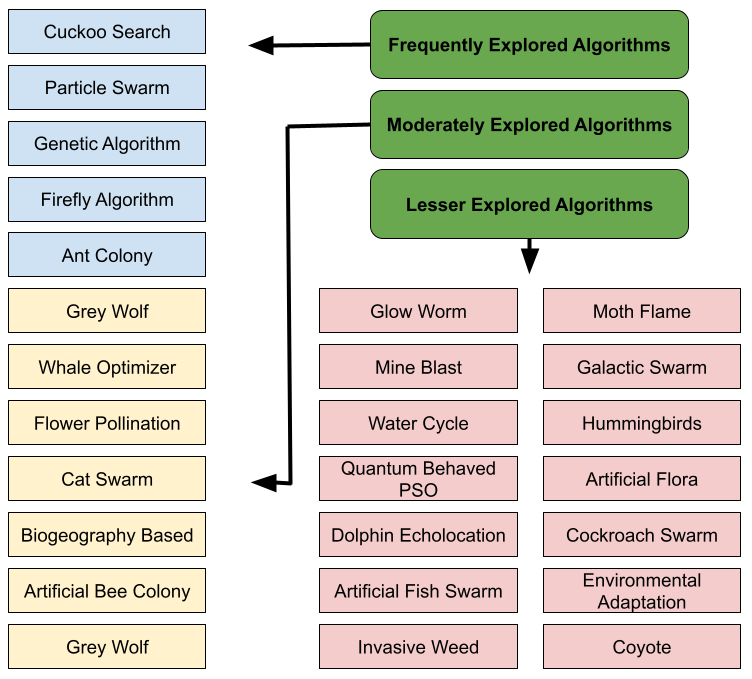}
    \caption{All nature-inspired meta-heuristic algorithms that were discussed in this paper divided into frequently explored, moderately explored and unexplored category}
    \label{fig:algorithms}
\end{figure*}

The papers within the year 2012-2024 were considered for review. Only peer-reviewed papers were considered from different journals, conference proceedings and books. The subject area of the survey constitutes Computer Science and Engineering. Only papers in the English language were surveyed. The papers were first collected based on the search terms and after removing duplicate ones, only 51 papers were left. After this, the title and abstract of all of these papers were thoroughly studied to eliminate papers irrelevant to the study area. The left-out papers were deeply studied to ensure their relevancy to the survey. Finally, out of these, the 28 most relevant papers were surveyed in this systematic review. Furthermore, during our exploration, we found a total of 26 nature-inspired meta-heuristic optimization algorithms which we have classified into three parts. The first part carries information about the 5 most commonly used algorithms. The second part includes the 7 algorithms which were explored in the state of the art but have further scopes for improvement. The third part contains a brief summary of the other 14 algorithms which were not explored for their applicability in the field of biomedical image and signal processing. Figure ~\ref{fig:algorithms} shows the division of frequently explored, moderately explored and lesser explored algorithms.

\section{Frequently Used Algorithms}
\label{sec:freq-algorithms}
This section discusses the algorithms which were thoroughly explored, texted and found to be reliable.

\subsection{Cuckoo search}
The Cuckoo search algorithm is a meta-heuristic algorithm inspired by the brood parasitism of some species of Cuckoo bird~\cite{YANG2014129}. Cuckoos lay eggs in nests of some other host birds of a different specie. Sometimes, they may even remove the eggs of the host bird to increase the chances of survivability of their own eggs. If the host bird discovers that there are foreign eggs in their nest, it either gets rid of these foreign eggs or abandons the nest and builds a new one. Recent studies show that the cuckoo search has been found to perform better than many other nature-inspired algorithms like particle swarm optimizer, genetic algorithm, etc. in many cases. The algorithm is based on the following rules:

\begin{itemize}
    \item one egg is laid by each cuckoo at a time and then it dumps the egg in a randomly chosen nest.

    \item the nests containing the best eggs are carried to the next iteration.

    \item the number of nests available is set. The probability of discovering the cuckoo eggs by the host bird is $p_a \in (0,1)$.

    \item if the eggs are discovered, the host bird can either get rid of the egg or build a new nest.
\end{itemize}

This algorithm can be extended as a deep-learning stochastic optimizer. For that, firstly a new set of solutions (eggs) are generated using Levy's flight method. For a weight matrix $x$, at $(t+1)^{th}$ iteration, learning rate $\alpha$, Heviside function $H(u)$, random number $\epsilon$,

\begin{equation}
    x_i^{(t+1)} = x_i^t + \alpha\otimes L(\lambda)
\end{equation}
where,
\begin{equation}
    L(s,\lambda)=\frac{\lambda\Gamma(\lambda)sin(\pi\lambda/2))}{\pi} \times \frac{1}{s^{(1+\lambda)}}, (s >> s_0 > 0)
\end{equation}
here, step size is $s$.

The algorithm uses a combination of local and global random walk. The local walk is written as,
\begin{equation}
    x_i^{(t+1)} = x_i^{(t)} + \alpha s \otimes H(p_a-\epsilon)\otimes(x_j^t-x_k^t)
\end{equation}
Here, $x_j^t$ and $-x_k^t$ are randomly selected solutions. And the global random walk is performed using L\'evy flight method as given by,
\begin{equation}
    x_i^{(t+1)} = x_i^{(t)} + \alpha L (s,\lambda)
\end{equation}

Therefore, in this way, the weights and biases of the neural network are updated in each iteration to find the global minima. Cuckoo search works extremely fast to find the local minima and therefore it keeps looking for other local minima and the best fit is considered the global minima.

In 2023, ~\cite{guo2013novel} proposed a method for ECG signal compression by application of DWT and several nature-inspired optimization algorithms. This includes particle swarm optimization, artificial bee colony optimization, cuckoo search algorithm, and flower pollination algorithm. Among all the algorithms, cuckoo search was found to perform better for ECG signal compression. In 2022, ~\cite{doi:10.1089/cmb.2021.0410} published an article showing the application of cuckoo search optimization for the detection of cancer based on microarray data. Gene was extracted from the data with independent component analysis and the classification was performed with Na\"{i}ve Bayes classifier. Cuckoo search and artificial bee colony hybrid were used for feature extraction and dimensionality reduction. The cuckoo search improved the local search process of the artificial bee colony algorithm. The results from the experiment show that this method along with independent component analysis performs a deeper search for iterative processes. This can be used to avoid convergence to local optima. In 2022, ~\cite{Das2022} have proposed a method for ECG signal denoising by using a orthogonal wavelet filtering method whose coefficients have been optimized with cuckoo search algorithm. The method was verified on PhysioBank database. The comparison of the results with empirical wavelet transform, empirical mode decomposition and Butterworth high-pass filtering methods have shown satisfactory performances for removal of various ECG signal noises even for low input signal-to-noise-ratio. In 2022, ~\cite{Aziz20221627} proposed a mechanism for informative gene selection from a gene expression dataset. The cuckoo search algorithm was used in this case for dimension reductionality and maximizing classification accuracy. The exploitation phase in this algorithm was accompanied by artificial bee colony algorithm and the exploration phase was accompanied by genetic algorithm. Also, independent component analysis was used for feature extraction. Later Na\"{i}ve Bayes classifier was used to classify different classes accociated with gene expression. The proposed method was tested on 6 different benchmark datasets. The results from the experiment shows that the method can be effective in avoiding premature convergence and produce better performance compared to the state of the art. In 2017, ~\cite{8169893} have propsed a novel extension of binary cuckoo search that can be used for feature selection. The paper uses a pseudobinary mutation neighbourhood search to improve the state-of-the-art cuckoo search. To test the performances of this modification, the algorithm have been tested on several biomedical datasets. Performances on these datasets were compared against other meta-heuristic algorithms like ant colony, genetic algorithm and particle swarm. The tests verifies that the proposed modification of cuckoo search outperforms all other mentioned optimizers in terms of accuracy.

\subsection{Particle Swarm Optimization Algorithm}
Particle Swarm optimizer (PSO) is one of the most popular nature-inspired algorithms introduced in 1995 by Kennedy and Eberhart. It has reputation for faster and cheaper convergence compared to other classical optimizers and it can also be parallelized. Also, as it does not use the gradient of the objective function, it does not require the objective function to be differentiable. The algorithm is inspired by the social behaviour of a group of fish or birds. The group of birds works together to find their prey. When one of the bird finds prey while a random search, all the other birds in the flocks are notified and all of them starts to chase the target and eventually obtains a quicker hunt. The PSO algorithm works in a similar fashion where the weights of the neural network model are independently seeking for the global minima until one of the weights reaches an approximate minima and then all the other weights try to chase those minima aiming for faster convergence. The particles (or weights for a neural network) initially do not know where the global minima are located. But using the fitness function $P$, the fitness values of the particles can be evaluated.

\begin{equation}
    P^t_i = [x^t_{0,i}, x^t_{1,i}, x^t_{2,i}, ... , x^t_{n,i},]
\end{equation}

As the particles moving through the search space have a velocity, therefore the velocity vector of the particles is defined by,

\begin{equation}
    V^t_i = [v^t_{0,i}, v^t_{1,i}, v^t_{2,i}, ... , v^t_{n,i},]
\end{equation}

The weight vector and velocity vector are initialized with random values. The goal of each weight is to reach optima sooner than others. Each particle is accelerated stochastically towards the global best solution. The weight update rule is given by,

\begin{equation}
    P^{(t+1)}_i = P^{(t)}_i + V^{(t+1)}_i
\end{equation}

where,
\begin{equation}
    V_i^{(t+1)} = wV_i^{(t)} + c_1r_1(P^{t}_{best(i)}-P^{(t)}_i) + c_2r_2(P_{global\_best}^{(t)}-P^{(t)}_i)
\end{equation}

In 2016, ~\cite{OLIVEIRA2016366} proposed a process for integrating artificial intelligence into a feedback control topic addressing mechanism. This mechanism works by tuning of proportional, integrative and derivative (PID) controllers. The work was conducted in an introductory undergraduate course. Two biomedical problems were addressed in this, first is the arterial pressure control and another is minimum temperature control for the treatment of intracranial tumours. The particle swarm optimizer was used here to optimize the control system of the PID controllers. The results from the experiment show that the method helped the students to better understand the design conflicts arising from set-point tracking and load disturbance rejection. The results further confirmed that PSO has optimized the performances of PID for 2 degrees of freedom. In 2020, ~\cite{Bilandi20211410} proposed a framework by simulated annealing algorithm based on a hybrid particle swarm optimization for relay node selection for its application in wireless body area network (WBAN). The paper had shown that the optimizer can be used to make optimal WBAN design for improving energy efficiency. The proposed model have shown to perform better than 12 other metaheuristic algorithms. In 2022, ~\cite{ncta22} have shown that particle swarm optimizer can be used for extraction of most informative and discriminative channels from multi-lead EEG signals. Later, self-evolving radial basis functions neural network model was used for motor imagery recognition. The proposed method was tested on two open sourced EEG dataset and the method was found to outperform many of the existing techniques for motor imagery classifications in terms of efficiency. In 2018, ~\cite{Bejinariu2018280} have proposed a method for biomedical image registration using different nature-inspired algorithms like particle swarm, multi-swarm and cuckoo search. The geometric transformation as obtained by these optimizers had seven parameters. This includes rotation against a point within the image. The results from the experiment based on 25 tests on each image registration procedure shows that particle swarm produces the most accurate solutions. However, cuckoo search and multi-swarm have less scattered solutions. The swarm techniques had higher convergence speed. In 2022, a paper was published ~\cite{9689046} which has demonstrated the usage of hybrid particle swarm and firefly optimizer-based method for embedding patient information in their biomedical images. The work was basically an application of steganography on biomedical images as a cover image and patient information as payloads. The proposed hybrid meta-heuristic algorithm was found to obtain a no-attack score of 0.9985 which confirms its high security.

\subsection{Genetic Algorithm}
Genetic algorithm (GA) is a very commonly used nature-inspired evolutionary algorithm which is used to solve a variety of problems including optimization problems. This is inspired by Darwin's evolution theory of Natural Selection. Chromosomes are the units within a cell that carries genomic information. When cells of a living being replicate, two chromosomes come together to exchange some of the genetic materials (genes). The process is called crossover. This results in offspring with some properties from one chromosome and some from the other. This is the primary cause for variations in the genome of the offspring. The other causes of variation include mutation which is the error introduced while copying the genetic material. The process is mathematically simulated for solving optimization problems. Firstly, a population is randomly generated whose size depends upon the nature of the problem. The next step is the parent selection. A fitness function is used to measure the fitness of the solution. At this point, the parent could be selected either randomly, by taking one from every group of subpopulations (depending on the nature of the problem), or by using a probabilistic method called roulette wheel selection. After the selection of parents, a crossover is performed between the parents to produce offspring. A mutation rate can be affixed to mutate the properties of the offspring at a given rate. The fitness of the offspring is evaluated. The best offspring are merged with the main population and the remaining are discarded. The process is kept repeating over a number of generations until an acceptable fit is obtained within a generation.

In 2018, ~\cite{8324449} proposed a model for developing bio-inspired soft robotic systems and their mass production. For achieving a gradient-free design optimization technique, Genetic Algorithm was used. The model has been presented through a case study where the authors have built a bio-inspired soft pneumatic actuator. The results from the experiment show that a larger population of chromosomes lead to a faster convergence taking lesser resources than the usual stochastic uniform selection method. ~\cite{9596077} in 2021 proposed a model for the classification of Electromyogram (EMG) signals. The authors have used different signal processing techniques for noise reduction and then used the genetic algorithm to perform dimensionality reduction. The transformed model is then processed through classification models and for the Electro-Myography EMG dataset, the classification accuracy was obtained to be 95\%. In 2015, ~\cite{https://doi.org/10.1049/iet-smt.2015.0048} proposed a method for sub-band adaptive filtering technique which can be used to denoise EEG data. The algorithm was attempted to be built with 5 evolutionary techniques like particle swarm, artificial bee colony, cuckoo search, genetic algorithm and differential evolution. Among all these, the genetic algorithm was found to be the best. This was because, for GA, the controlling of the parameters like random mutation and random selection of chromosomes could be controlled within a predefined range which cannot be done with other algorithms. In 2019, ~\cite{Zuvela20198101} have shown a method for in vivo diagnosis and detection of nasopharyngeal carcinoma from the Raman spectroscopy technique by using genetic algorithm. The algorithm was tested on a total of 2126 Raman Spectra of which 598 samples were nasopharyngeal carcinoma positive. The data were obtained from 113 tissue sites of 14 nasopharyngeal carcinoma-positive subjects and 48 healthy subjects. The results from the experiment indicated a 98.23\% detection accuracy and show scopes for its implementation in real-time in vivo spectroscopy-based nasopharyngeal carcinoma detection.

\subsection{Firefly algorithm}
The Firefly algorithm was developed inspired by the light works of fireflies for attracting potential mates. Fireflies with brighter lights tend to attract mates more than other fireflies. Also, the distance between them inversely impacts the attraction. This behaviour is mathematically modelled for optimization. The assumptions for this modelling are:
\begin{itemize}
    \item All fireflies in the swarm are unisexual indicating any firefly can be attracted to any other firefly in the swarm depending on other criteria.

    \item Attractiveness of the fireflies is directly proportional to the brightness of the firefly and inversely proportional to the distance between them.

    \item Finally, the objective function indicates the brightness of the firefly.
\end{itemize}

The first step for this is to initialize the objective function $f(x_i)$. From the inverse square law of light, we know that for distance $r$, intensity $I(r)$ varies by,
\begin{equation}
    I(r) = \frac{I_s}{r^2}
\end{equation}
Considering absorption coefficient $\gamma$,
\begin{equation}
\label{eqn:brithness-firefy}
    I = I_0e^{\gamma r^2}
\end{equation}
The initial population $n$ of fireflies are initialized by,
\begin{equation}
    x_{t+1} = x_t+\beta_0e^{-\gamma r^2} + \alpha\epsilon
\end{equation}
where the second term arises by the attraction and the third term is for randomization where $\alpha$ is the randomization parameter.

After this, calculate the light intensity or brightness of each firefly by eqn. ~\ref{eqn:brithness-firefy}. Then based on this, evaluate the attractiveness of each firefly by,
\begin{equation}
    \beta = \beta_0 e^{-\gamma r^2}
\end{equation}
Now, the fireflies move towards the brighter or more attractive firefly. This movement for less attractive firefly $i$ towards more attractive firefly $j$ is determined by,
\begin{equation}
    x_i = x_i + \beta_0 e^{-\gamma r_{i,j}}(x_j-x_i)+\alpha\epsilon
\end{equation}
Following this step, update the light intensities of the fireflies and rank them to find the current best solution.

In 2014, a work conducted by ~\cite{dey2014firefly} have shown the usage of the firefly algorithm to optimize the scaling factors for embedding medical information in ophthalmologic imaging. The goal of the work is to build a biomedical content authentication system by embedding different information like hospital logos and electronic patient records into the retinal image of the patient utilization of a combination of discrete wavelet transformation, discrete cosine transformation and singular value decomposition methods. The firefly algorithm was used here for finding the best scaling factors for embedding. The algorithm was proven to be an accurate authentication system for information exchange. Another work from 2020, ~\cite{9299273}, was conducted by application of the firefly algorithm on EEG data to perform feature selection in order to classify the EEG signal. The test results on an open-sourced dataset consisting of data from 29 subjects show that the average classification accuracy was obtained to be 76.14\% with the k-nearest neighbour classification algorithm. This was found to be a 4.4\% improvement from raw data classification.

\subsection{Ant colony optimization}
Ant Colony Optimization (ACO) was introduced in the 90s by Marco Dorigo in his PhD thesis. The algorithm was inspired by the foraging behaviour of ants for seeking paths between their food source and colony. Earlier, the algorithm was used to solve travelling salesman problems but later it was used to solve hard optimization problems as well. Ants live in colonies and hop from one place to any to search for food. While moving, they keep dropping pheromones on the ground which is an organic compound to mark the territory, draw the path, communicate with other ants, etc. After an ant finds some food, it gathers as much it can carry and while moving towards its colony keeps dropping pheromones on the path based on the quantity and quality of the food available at the source. Other ants that smell the pheromone follow that path. The more the pheromone in a path, the chances are more ants would follow that path. And more ants follow that path, more pheromones are deposited on the trail. The behaviour is mathematically modelled. For this, each ant is denoted by $k$ which is used to compute a set $A_k(x)$ of feasible current state expansions during each iteration. The ants move from state $x$ to $y$ in each iteration. The probability $p_{xy}^k$ for each ant for moving from state $x$ to $y$ depends on two factors. First is the attractiveness $\eta_{xy}$ of the move and the trail level $\tau_{xy}$. The probability is given by,
\begin{equation}
    p_{xy}^k=\frac{(\tau_{xy}^\alpha)(\eta_{xy}^\beta)}{\sum_{z\in allowed_x}(\tau_{xz}\alpha)(\eta_{xz}^\beta)}
\end{equation}
Here, the deposited pheromone volume is given by $\tau_{xy}$ while moving from $x$ to $y$. The control influence for $\tau_{xy}$,$\eta_{xy}$ is a parameter $0\leq\alpha$. $\beta\geq1$ is a control influence parameter of $\eta_{xy}$, $\tau_{xz}$ and $\eta_{xz}$.

After all the ants have completed their solution, the trails are updated. The pheromone update rule is given by,
\begin{equation}
    \tau_{xy}\longleftarrow(1-\rho)\tau_{xy}+\sum_{k}^{m}\Delta\tau_{xy}^k
\end{equation}
Where $\rho$ denotes the pheromone evaporation coefficient and $m$ denotes the number of ants. For cost $L_k$ for the tour of $k^{th}$ ant, constant Q, the amount of pheromone deposited $\Delta\tau_{xy}^k$ is given by,
\begin{equation}
\Delta\tau_{xy}^k =
\begin{cases}
Q/L_k, \text{~if ant $k$ follows $xy$ in its tour} \\
0, \text{otherwise}
\end{cases}
\end{equation}

In 2022, ~\cite{bangotra2022energy} published an article for building an optimistic routing protocol for wireless sensor networks. The system is intended to be energy-efficient. The framework is built for its application in facilitating smart healthcare systems. The results from the experiment show that the trust-based secure intelligence opportunistic routing protocol algorithm was found to perform better than many swarm-based nature-inspired algorithms like particle swarm, ant colony optimization and bacterial foraging. It was further found that nature-inspired algorithms can be used to find the best routes with modifications. Both methods have been used in a hybrid fashion for creating the fastest route finder. In 2015, a paper published ~\cite{7391459} tested a total of 72 classifiers which had a blend of nature-inspired meta-heuristic approaches and the classifiers were tested on 29 test cases based on 10-fold cross-validation. The study concluded that the Ant Colony Optimization hybrid with Decision Tree to build a classification algorithm performs significantly better (alpha=0.5) than the other 71 classifiers in 32 datasets compared to 41 distinct classifier instances. In 2022, ~\cite{9730603} proposed a method for the compression of ECG data using a tunable-Q wavelet transform method. The parameters of this wavelet transform method were optimized by using the ant colony optimization method. The test results from the experiment show that the compression ratio was found to be 22.42, the percentage root mean square difference was found to be 4.52\% and a quality score of 6.05 was achieved.

\textbf{\textit{Summary--}} Out of all the nature-inspired meta-heuristic algorithms, the cuckoo search was found to be the most reliable and widely tested. Particle swarm, genetic algorithm, firefly algorithm and ant colony optimization algorithms are also very widely tested in different optimization tasks like signal noise reduction, signal compression, feature selection and more. All of these algorithms can be safely implemented in the widest range of biomedical applications.

\section{Moderately Explored Algorithms}
\label{sec:moderately-algorithms}
This section discusses all the algorithms which were moderately explored.

\subsection{Grey Wolf Optimizer}
Grey Wolf Optimization (GWO) is another nature-inspired meta-heuristic optimization algorithm proven to work better than PSO in multiple instances. It is based on the social and hunting behaviour of grey wolves. Alpha ($\alpha$) males are the leaders of the pack and are the decision maker. Beta ($\beta$) males are subordinates of alpha who convey the order of the alpha to the remaining pack. Delta $\delta$ are below the alpha and beta but are higher in the social ranking than omega $\omega$. Omega are the least important members in the pack who are only allowed to eat at last. The grey wolves hunt in two phases, one is exploration and another is exploitation. For coefficient vectors $\vec{A}$ and $\vec{C}$, the position vector of the prey $\vec{X_p}$, and the position vector of a grey wolf $\vec{X}$,
\begin{equation}
    \vec{D} = |\vec{C}.\vec{X_p}(t)-\vec{X_p}(t)|
\end{equation}
\begin{equation}
    \vec{X}(t+1) = \vec{X_p}(t)-\vec{A}.\vec{D}
\end{equation}
where, $t$ indicates the current iteration. Therefore, the next iteration is given by,
\begin{equation}
    \vec{X_1}(t+1) = X(t)-\vec{A_1}.\vec{D}
\end{equation}
and,
\begin{equation}
    \vec{A} = 2\vec{a}\vec{r_1}-\vec{a}
\end{equation}
\begin{equation}
    \vec{C} = 2\vec{r_2}
\end{equation}
where $\vec{r_1}$ and $\vec{r_2}$ and the components of $\vec{a}$ linearly decreases from 2 to 0 in consequent iterations.

During each iteration, according to the position of the alpha, beta and delta, the omega wolves update their position as they have better knowledge about the potential location of the prey. This sequence of steps is given by,
\begin{equation}
    D = |\vec{C_1}.\vec{X}(t)-\vec{X}(t)|,
    D = |\vec{C_2}.\vec{X}(t)-\vec{X}(t)|,
    D = |\vec{C_3}.\vec{X}(t)-\vec{X}(t)|
\end{equation}
\begin{equation}
    \vec{X_1}(t+1) = \vec{X}(t)-\vec{A_1}.\vec{D}
\end{equation}
\begin{equation}
    \vec{X_2}(t+1) = \vec{X}(t)-\vec{A_2}.\vec{D} 
\end{equation}
\begin{equation}
    \vec{X_3}(t+1) = \vec{X}(t)-\vec{A_3}.\vec{D} 
\end{equation}
from these,
\begin{equation}
    \vec{X}(t+1) = (\vec{X_1}+\vec{X_2}+\vec{X_3})/3
\end{equation}

This completes the exploration phase where the grey wolves find the potential location of the prey. Following this, the exploitation or the attack phase begins when the grey wolves ultimately attack the prey or here in our case, hits the global optima. This is done by decreasing $\vec{a}$. And $\vec{A}$ is any arbitrary value in $[-2a,2a]$ where $a$ is decreased from 2 to 0 in subsequent iterations. For, $|\vec{A}|<1$, the wolves attack the prey and for $|\vec{A}|>1$ the grey wolves diverge from the current prey (optima) and continue to look for better prey.

In 2019, ~\cite{8711735} published a method for the usage of a grey wolf optimizer for building an adaptive noise cancellation (ANC) algorithm. The algorithm has been tested on Electroencephalogram (EEG) and Event-Related Potentials (ERP). The noise cancellation algorithm also uses gradient-based and swarm-based algorithms. The ANC is built by using GWO to update the weights to an optimal value. The experiment has been verified by introducing White Gaussian Noise (WGN) to EEG signals at various signal-to-noise ratio (SNR) levels. The performance shows that the GWO produces better results than PSO and other gradient-based methods.

\subsection{Whale optimizer algorithm}
The whale optimization algorithm (WOA) is another meta-heuristic method proposed by Mirjalili (same author as GWO) in 2016. This is based on the feeding behaviour of a humpback whale known as bubble-net hunting strategy. Humpback whales prefer to hunk school of krill or small fishes near the surface. They start hunting by creating a distinctive bubble net along a circle, spiral or in the shape of a '9'. Here the humpback whales dive around 12m deep, create a bubble net around the prey and swim upwards. The fish are trapped inside this bubble net, and later the whale feeds on these. This behaviour is mathematically simulated for the whale optimization algorithm. In this simulation, initially, the best solution is assumed to be the closest to the prey and the remaining solutions update their position towards the current best solution. For coefficient vectors $\vec{A}$ and $\vec{C}$, position vector $\vec{X}_{best}$ of best solution, and position of the whale $\vec{X}$,

\begin{equation}
    \vec{D} = |\vec{C}.\vec{X_{best}}(t)-\vec{X}(t)|
\end{equation}
\begin{equation}
    \vec{X}_{(t+1)} = \vec{X_{best}}(t)-\vec{A}.\vec{D}
\end{equation}
where,
\begin{equation}
    \vec{A} = 2\vec{a}\vec{r_1}-\vec{a}
\end{equation}
\begin{equation}
    \vec{C} = 2\vec{r_2}
\end{equation}
where $\vec{r_1}$ and $\vec{r_2}$ are random vectors within [0,1] range.

The simulation of the bubble-net shrinking encircling mechanism is achieved by a gradual reduction of the value of $\vec{a}$ from 2 to 0 during subsequent iterations.

Now, the spiral position update is given by,
\begin{equation}
    \vec{D'} = |\vec{X}_{best}(t)-\vec{X}(t)|
\end{equation}
\begin{equation}
    \vec{X}(t+1) = \vec{D'}.e^{bl}.cos(2\pi l)+\vec{X}_{best}(t)
\end{equation}
where $l$ is a random number within [-1,1] range. 

The humpback whales randomly search for prey according to their position of each other. This behaviour is given by,
\begin{equation}
    \vec{D} = |\vec{C}.\vec{X}_{random}(t)-\vec{X}(t)|
\end{equation}
\begin{equation}
    \vec{X}{(t+1)} = \vec{X}_{random}(t)-\vec{A}.\vec{D}
\end{equation}

In 2019, ~\cite{doi:10.4015/S1016237219500352} have shown a method for adaptive noise cancellation (ANC) method for denoising EEG and ERP signals by application of Oppositional Whale Optimization Algorithm (OWOA). The ANC performances are improved by updating the weights by the OWOA algorithm. White Gaussian Noise have been introduced to signals at various signal-to-noise ratio levels like (-10 dB, -15dB and -20dB). The results have been compared with methods like recursive least square, least mean square, genetic algorithm, particle swarm optimization and general whale optimization algorithm. The results from the experiment confirm that the proposed method has better signal-denoising capabilities compared to the state-of-the-art.

\subsection{Flower pollination algorithm}
    In 2018, ~\cite{SANJOSEREVUELTA201892} conducted a workshop for building a frequency-selective finite impulse response filter based on a flower pollination algorithm. The algorithm was tested on EEG signals. The results from the experiment show that the proposed method has obtained 5-10 times larger attenuation in the stop band and a narrower transition band. This has been achieved with the expense of increasing the passband ripple within the 5-15\% range in 75\% of the cases. In 2022, a work ~\cite{doi:10.1080/03772063.2022.2088627} was published which demonstrated the usage of a flower pollination algorithm for congestive heart failure detection based on ECG signals. The work tested different variants of the flower pollination algorithm on ECG data for feature selection. Later used a support vector machine with various kernels, k-nearest neighbours and Na\"{i}ve Bayes classifier. The experimental results show the highest detection accuracy of 97.71\%. In 2020, ~\cite{9007451} showed a method for schizophrenia detection from EEG signals by using the flower pollination algorithm. The experiment used techniques like expectation maximization integrated with partial least squares, non-linear regression and partial least squares for feature extraction. Optimizers like flower pollination, eagle strategy with the help of differential evolution, backtracking search and group serach optimization algorithms were used to further optimize the extracted features. Later, Adaboost and Na\"{i}ve Bayes classifiers were used for classification. The flower pollination based algorithm was found to achieve a classification accuracy of 98.77\%.

\subsection{Cat swarm optimization}
Next on the list is Cat Swarm Optimization (CSO). This was proposed in 2006 and was inspired by the resting and tracing behaviour of cats. Cats apparently remain lazy spending most of their time resting. But, during this resting phase, they stay highly alert to what happening around them. Whenever they notice prey, they start to move towards it quickly. This behaviour is mathematically modelled for solving the optimization problem. The two manoeuvres associated with this behaviour is tracing and seeking. Each cat here represents a solution set consisting of position, fitness value and a flag. A few terms associated with this technique are Seeking memory pool (SMP), self-position considering (SPC), counts of dimension to change (CDC) and seeking a range of selected dimensions (SRD)~\cite{ahmed2020cat}. The seeking consists of the following phases:
\begin{itemize}
    \item Make a large number of SMP copies of the current position of the cat.

    \item For every copy of the SMP, make a random selection of CDC to be mutated. Also, randomly add or subtract SRD values from the currently available values replacing the older positions. This is given by:

    \begin{equation}
        Xjd_{new} = (1 + random \times SRD) \times Xjd_{old}
    \end{equation}
    where, $Xjd_{old}$ is the previous position, $Xjd_{new}$ is the next position. The number of cats and dimensions are respectively given by $j$ and $d$. Here, $random$ is any arbitrary number belonging to [0,1].

    \item After this, evaluate the fitness $F$ of all the candidate positions.

    \item Finally, depending on probability $P_i$ based on fitness values, select the next position for the cat. If all fitness values are equal, then set it as 1.
    \begin{equation}
        P_i = \frac{|F_i-F_b|}{F_{max}-F_{min}}, where~0<i<j
    \end{equation}
    for minimization, $F_b=F_{max}$ and for maximization, $F_b=F_{min}$.
\end{itemize}

Following this, the tracing motion of the cat is simulated. The position vector of all cats is randomly initialized during the first iteration. The steps undergone for this are as follows:

\begin{itemize}
    \item The velocities $(V_{k,d})$ for all dimensions are updated by,
    \begin{equation}
        V_{k,d} = V_{k,d} + r_1c_1(X_{best,d}-X_{k,d})
    \end{equation}
    
    \item The position of $Cat_k$ are consequently updated by,
    \begin{equation}
        X_{k,d} = X_{k,d} + V_{k,d}
    \end{equation}
\end{itemize}

\subsection{Grasshopper optimization algorithm}
Grasshoppers are the insects that form the world's largest swarm, sometimes spreading up to continental range. They are very destructive and do large-scale damage to crops. The life cycle consists of three phases which are eggs, nymph and adulthood. The swarms are formed right from the nymph stage and during this stage due to lack of wings, they cover small distances by hopping. But during adulthood, they have wings and they cover large distances~\cite{SAREMI201730}. This swarming behaviour of grasshoppers is mathematically modelled to solve optimization problems. The characteristics of grasshoppers considered for modelling are:
\begin{itemize}
    \item During the nymph stage, the swarm moves slowly as they do not have wings.
    \item During adulthood, the swarm moves large distances as they have wings.
    \item The swarm performs the food search in two stages namely exploration and exploitation.
\end{itemize}
For social interaction, $S_i$ between the grasshoppers and the solution, the gravity force on the solution $G_i$ and wind advection $A_i$, the position of solutions $X_i$ is given by,
\begin{equation}
\label{eqn:grass-position}
    X_i = S_i + G_i + A_i
\end{equation}
For random numbers $r1$, $r2$ and $r3$ in range [0,1],
\begin{equation}
    X_i = r_1S_i + r_2G_i + r_3A_i
\end{equation}
The force of social interaction is defined by,
\begin{equation}
\label{eqn:grass-social}
    S_i = \sum_{j=1}^Ns(d_{ij})\hat{d_{ij}},~where~i\neq j
\end{equation}
\begin{equation}
\label{eqn:grass-social-forces}
    s=fe^{\frac{-r}{l}}-e^{-r}
\end{equation}
where, $d_{ij}=|x_j-x_i|$ denotes the distance between two grasshoppers, $\hat{d_{ij}}=\frac{|x_j-x_i|}{d_{i,j}}$. Here, $s$ represents attraction and repulsion between grasshoppers, the scale of attractive length is given by $l$ and the intensity of attraction is given by $f$.

The force of gravity $G_i$ here is calculated by,
\begin{equation}
\label{eqn:grass-gravity}
    G_i = -g\hat{e_g}
\end{equation}
where $\hat{e_g}$ directs towards the center of the earth and $-g$ is the gravitational constant.

The wind direction $A_i$ is measured by,
\begin{equation}
\label{eqn:grass-wind}
    A_i = u\hat{e_w}
\end{equation}
where $\hat{e_w}$ directs towards the wind direction and $u$ denotes the drift constant.

For the grasshopper position, eqn. ~\ref{eqn:grass-position} can be modified using eqn. ~\ref{eqn:grass-social}, \ref{eqn:grass-social-forces}, \ref{eqn:grass-gravity} and \ref{eqn:grass-wind}.
\begin{multline}
\label{eqn:grass-full}
    X_i = \sum_{j=1}^{N}s(d_ij)\hat{d_{ij}}-g\hat{e_g}+u\hat{e_w} \\
    = \sum_{j=1}^{N}s(|x_j-x_i|)\frac{|x_j-x_i|}{d_{ij}}-g\hat{e_g}+u\hat{e_w},~ where i\neq j
\end{multline}
To avoid local optima, eqn. ~\ref{eqn:grass-full} is further modified.
\begin{multline}
    X_i^d = c(\sum_{j=1}^{N}c\frac{UB_d-LB_d}{2}s(|x_j^d-x_i^d|)\frac{|x_j-x_i|}{d_{ij}} \\
    + Best~Solution,~where~i\neq j
\end{multline}
where $G=0$, $d^{th}$ dimensional best solution is given by $A$, $UB_d$ and $LB_d$ are the upper and lower bounds for the $d^{th}$ dimension. Here, $c$ is a coefficient which decreases with each passing iteration. For $c_{max}$ and $c_{min}$ respectively indicating the maximum and minimum value of $c$, $iter$ denoting the current iteration and $Max_{iter}$ denoting the maximum allowed iteration, $c$ is given by,
\begin{equation}
    c = c_{max}-iter\frac{c_{max}-c_{min}}{Max_{iter}}
\end{equation}

In ~\cite{10.1007/978-3-319-74690-6_9}, authors have shown a hybrid grasshopper swarm optimization algorithm for automatized seizure detection from EEG signals. The authors have used a support vector machine with a radial basis function for the classification. The grasshopper swarm optimizer algorithm here was used for feature selection in order to reduce the dimensional complexity of the data and improve the classification performance.

\subsection{Biogeography-based optimization}
The biogeography-based optimization algorithm was proposed in 2008. It was inspired by the relationship of biological species for geographical migration. Immigration is inversely proportional to the number of species in a given area as with increasing species in an area, the survivability changes of the immigrants decrease. Now more emigration, the number of species is directly proportional to the emigration rate as with increasing species, it is easier for the species to emigrate or leave the region~\cite{harshavardhan2022lsgdm}.
In 2020, ~\cite{CHEN2020106335} proposed a method for using biogeography-based optimization techniques for biomedical image restoration. The work was conducted by using the Elite Learning method. 6 kinds of biomedical images obtained from 18 subjects were used to test the model. This makes 54 multimodel image registration scenarios out of which the proposed model performed best in 30 scenarios where the state-of-the-art methods obtained 21.

\subsection{Artificial bee colony algorithm}
An artificial bee colony is a very popular swarming algorithm. However, it has not been explored much for solving optimization problems in biomedical image and signal processing tasks. The algorithm mimicked the foraging behaviours of honey bees and was first introduced in 2005. The honey bees form a swarm and survive through social cooperation strategy. The primary objective of the queen bees and drone bees is to reproduce. And the foraging is performed by three types of bees namely worker bees, onlooker bees and scout bees. Now worker bees generally collect nectar from flowers and transfer it to the hive. They memorize the properties of the food sources and share the information with the onlooker bees. The onlooker bees are responsible for deciding the quality (fitness) of the food sources and instruct the worker bees to forage from a good food source. This phase comes to the exploitation phase. Some of the worker bees later act as scout bees whose primary work is to look for new and better food sources and this forms part of the exploration phase~\cite{OZTURK2020106799}.

\textbf{\textit{Summary--}} The 7 algorithms discussed here are very popular, reliable and well-tested in various application areas. However, their reliability was not fully established in state-of-the-art biomedical signal and image processing. Some of these algorithms have been tested in only a couple of works. The remaining algorithms have been tested in multiple different experiments but the results were not satisfactory. This leaves further scope for improvement for the given set of algorithms for their implementation in the biomedical image or signal processing applications.

\section{Less Explored Algorithms}
\label{sec:less-algorithms}
This section discusses the potential algorithms which were not explored.

\begin{enumerate}
    \item \textit{Glow worm optimization--}
    Glow-worm optimization has been inspired by the foraging behaviour of glow worms. They also attract partners through the glow. Luciferin is a light-emitting compound. Each glow worm is attracted by its neighbouring glow worm with higher luciferin. The algorithm works in three phases. Luciferin update, movement phase and local decision range update phase. Initially, all glow worms are randomly distributed in the solution space. During each iteration, the values are updated according to the current position of the glow worms according to the objective function. To simulate the luciferin decay over time, a fraction of it is subtracted during each iteration. After this, a glow worm chooses an adjacent glow worm having a higher luciferin value with probabilistic methods. During each iteration, the local decision range changes. Now, a glow worm having more luciferin attracts more glow worms and the higher density of the swarm indicates a larger local decision range~\cite{CHEN2017104}.
    
    \item \textit{Mine Blast algorithm--}
    In 2013, Mine Blast Algorithm (MBA) was first proposed ~\cite{SADOLLAH20132592}. It was inspired by the mine bomb explosion in a minefield. The idea is that when a mine bomb is exploded, the shrapnel from the explosion would spread around and may hit other nearby mine bombs. Now, the number of casualties can be compared to evaluate the fitness of the solution explosion. Here in this algorithm, the movement speed and direction of the shrapnels from each explosion are controlled. The basic concept is that large casualties by each shrapnel from an explosion indicate a higher number of mine bombs present in that area. An explosion with the highest casualties is considered to be the optimal solution.

    \item \textit{Water cycle algorithm--}
    The Water Cycle Algorithm was first proposed in 2012, inspired by the natural water cycle~\cite{ESKANDAR2012151}. The water is frozen as ice from glaciers when melts or rain drops at a high altitude region, they start flowing as water streams. They keep flowing downwards and merge with other water streams or rivers. Now when two streams combine, according to the algorithm, they form a second-order stream. When two second-order streams combine, they form a third-order stream. This continues until they finally mix into the sea. The initial population in the algorithm is formed by raindrops. Only the best-accumulated raindrops reach the sea. Therefore, only the rivers with the maximum flow ultimately reach the sea and this serves as the solution to the problem.

    \item \textit{Quantum-behaved particle swarm optimization--}
    This was first proposed in 2004, inspired by quantum computation methods. The goal here is that each particle of the particle swarm algorithm is treated as a quantum particle. The state of each particle is denoted by the wave function $|\psi(x,t)|^2$ according to the Schr\"{o}dinger wave equation in space-time domain instead of position-momentum domain. In this algorithm, the position $x_i$ and velocity $v_i$ cannot be simultaneously measured accurately. This behaviour helps to overcome some of the limitations of a regular particle swarm and improve the optimization performance~\cite{rugveth2023sensitivity}.

    \item \textit{Dolphin echolocation algorithm--}
    Dolphin uses a technique called echolocation to locate their prey. They send a sonar sound and estimate the distance of their prey from the reflected sonar waves thus received. While hunting, dolphin hunt in groups for the division of labour by cooperation. For hunting larger prey, dolphin calls other dolphins for help by sonar. The dolphins closer to the prey track the movement of the prey and the dolphin farther are responsible for surrounding the prey and restricting its escape routes. During the hunt, the dolphins communicate with each other with sonar for sending and receiving updates about their location. This predatory behaviour of dolphins is mathematically simulated in this algorithm to solve optimization problems.

    \item \textit{Artificial fish swarm--}
    The artificial fish swarm algorithm is inspired by the swarming behaviour of fish while getting food. Fishes sense the area with high food concentration by their vision or their various sense organs. The fishes tend to move in groups. When a fish finds some food, the other fishes behind start to follow them and reach that location quickly. This indicates that a higher density of fish in an area would indicate a higher chance of finding food in that region. When a fish randomly swims in water, it tends to seek a group of fish to ensure better chances of its survival. This swarming behaviour of fish is simulated to solve the optimisation problems effectively. This is also a good algorithm for avoiding local convergence~\cite{neshat2014artificial}.

    \item \textit{Invasive weed optimization--}
    Invasive weed optimization was first introduced in 2006 which is a simple but effective algorithm to reach global convergence. For the mathematical simulation, firstly the primary population is initialized by the distribution of a limited number of seeds around the search space. The seeds grow into a flowering plant which eventually produces seeds depending on their fitness value. The spread of grains from the flowers decreases non-linearly during each consecutive iteration eliminating appropriate plants and promoting more fit plants. To ensure a constant number of herbs in the colony, the grasses with the worst fitness are eliminated if the number of grass exceeds the maximum number of grass supported in that colony. The steps are iterated till convergence is achieved and then the colony with minimum cost function is promoted~\cite{MISAGHI2019284}.

    \item \textit{Moth-flame optimization algorithm--}
    The moths are attracted to flames and they navigate towards the flame by navigating with the help of the moon. The moths usually travel at night and take help from the moon for navigation to follow a transverse path. They keep a crosswise inclination towards the moon at a constant angle and based on that they maintain their path. When the distance between the moth and flame is less, they tend to revolve around it in a helical path. This mechanism of transverse travel and helical motion is mathematically simulated to solve optimization problems~\cite{sahoo2023moth}.

    \item \textit{Galactic swarm optimization--}
    Inspired by the movement of stars, galaxies and other heavenly bodies due to gravity, the galactic swarm optimization algorithm was used. The algorithm uses a hybrid mechanism of two other nature-inspired algorithms. It uses an artificial bee colony algorithm for the exploration phase due to its better exploration capabilities and particle swarm optimization for the exploitation stage because of its better convergence properties~\cite{bernal2020fuzzy}.

    \item \textit{Hummingbirds optimization algorithm--}
    Hummingbirds first visit a food source and evaluate its properties depending on the nectar quality (solution vector) and nectar refilling rate (fitness value). Once it finds the source appropriate, it regularly visits the source to collect nectar and remembers the visiting timetables. Now, hummingbirds also share information about their food source with other hummingbirds. Later, all of the hummingbirds start to collect nectar from the best available food source (convergence). This property of hummingbirds for nectar collection is simulated to formulate the optimization algorithm~\cite{ZHAO2022114194}.

    \item \textit{Artificial flora optimization algorithm--}
    The artificial flora optimization algorithm mimics the natural migration patterns of different plants. The migration of seeds can be done by the plant itself (e.g. explosion of the ripened fruit spreading the seeds upto 5m distance) is known as autochory and the migration can be done by other means (air, water, animal-borne, etc.) is known as allochory. The optimization algorithm works in such a way that if the seeds are migrated to a more fertile region, it would ensure their survivability or else extinct. This way, the region with highest survivability would indicate a convergence point~\cite{bacanin2022artificial}.

    \item \textit{Cockroach swarm optimization--}
    The cockroach swarm optimization algorithm was inspired by the swarming behaviour of cockroaches and also the behaviour observed while looking for food or escaping from light. The initial solutions are generated randomly in the solution space. The three phases of this algorithm include chase-swarming, dispersing and ruthless behaviour. In the chase-swarming phase, the local best solutions are carried by each cockroach in a small swarm and the swarm moves towards the global optima. Dispersion is performed by the cockroaches to take random steps in the search space which ensures a better exploration. Lastly, ruthless behaviour is shown by the cockroaches during which the weaker cockroaches are eaten by stronger cockroaches when there is food scarcity. These steps are performed iteratively to converge to a global optimum~\cite{e19050213}.


    \item \textit{Environmental adaptation optimization--}
    Environmental adaptation optimization is a method which considers the adaptation mechanism for organisms in a new environment. For a population having plastic traits when exposed to a new environment which is not optimal for that specific phenotype, then the subspecies with higher fitness value would be promoted and would be evolved in the process. The process is iterated till convergence is achieved~\cite{mishra2022advanced}. In 2019, ~\cite{crabtree2018building} proposed a method for the prediction of minimum free energy in an RNA secondary structure with help of an environmental adaptation optimization algorithm. The proposed method takes lesser fitness evaluation and is lesser expensive compared to the state-of-the-art methods. The technique achieves high functional accuracy and consistency for solving the energy optimization problem for secondary RNA structure.
    
    \item \textit{Coyote optimization algorithm--}
    The coyote optimization algorithm was developed in 2018 inspired by the swarming and hunting behaviour of coyotes'. The algorithm differs from a grey wolf optimizer as unlike a grey wolf optimizer, the coyote optimizer even though it has an alpha leader, it does not require a social hierarchy and has a different structural setup. It also makes use of social information-sharing methods instead of only hunting~\cite{8477769}. ~\cite{vineeth2021performance} have conducted an experiment in 2020 to denoise biomedical images. The study has considered several nature-inspired meta-heuristic algorithms like particle swarm, artificial bee colony, firefly algorithm, cuckoo search, crow search, spotted hyena and coyote optimization. The experimental results verified that the coyote optimization algorithms have better denoising capabilities compared to other algorithms. However, artificial bee colony have been found to be the fastest to converge.
    
\end{enumerate}

\textbf{\textit{Summary--}} The 14 algorithms discussed here are not explored in the state-of-the-art biomedical signal and image processing. Although these algorithms have shown promising results in other application domains, they are yet to be tested in the mentioned domain. These leaves a knowledge gap and further scope for exploration.
\section{Conclusion}
\label{sec:conclusion}
In search for better optimization techniques, several nature-inspired meta-heuristic algorithms have been proposed in the state of the art. In the current literature, there are no peer-reviewed surveys published which have discussed the recent advancements of nature-inspired algorithms for application in biomedical signal and image processing. To solve this knowledge gap, the paper attempts to identify all recent relevant works and has reviewed 28 latest relevant papers. The study has identified 5 such algorithms which can be reliably used for various applications in the biomedical engineering domain including signal compression, signal denoising, feature selection and more. Further 7 algorithms were identified which were performing well in other application areas but their full potential was partially explored for applications in biomedical engineering. Moreover, 14 more promising algorithms were identified which were not explored in the state of the art.

\section*{Acknowledgement}
The work is a part of the Gyanam Foundation. The authors would like to thank Wingbotics LLP and Spiraldevs Automation Industries for their overall support.

\bibliographystyle{cas-model2-names}
\bibliography{cas-refs}



\end{document}